%% file: root.tex
\documentclass[letterpaper, 10 pt, conference]{ieeeconf}  %

\IEEEoverridecommandlockouts                              %

\usepackage{tabu}
\usepackage{blindtext}
\usepackage{graphicx}
\usepackage{xspace}
\usepackage{xcolor}
\usepackage{booktabs}
\usepackage{xspace}
\usepackage{multicol}
\usepackage{float}
\usepackage{multirow}
\usepackage{multirow}
\usepackage{xcolor} 
\usepackage{makecell}
\usepackage{tabu}
\usepackage{amsmath,amssymb,amsfonts}
\makeatletter
\let\NAT@parse\undefined
\makeatother
\usepackage{hyperref} 
\newcommand{\PreserveBackslash}[1]{\let\temp=\\#1\let\\=\temp}
\newcolumntype{C}[1]{>{\PreserveBackslash\centering}p{#1}}
\newcolumntype{R}[1]{>{\PreserveBackslash\raggedleft}p{#1}}
\newcolumntype{L}[1]{>{\PreserveBackslash\raggedright}p{#1}}
\usepackage{xspace}
\makeatletter
\DeclareRobustCommand\onedot{\futurelet\@let@token\@onedot}
\def\@onedot{\ifx\@let@token.\else.\null\fi\xspace}

\def\eg{\emph{e.g}\onedot}

\def\etc{\emph{etc}\onedot}

\makeatother

\input{shortcut}

\title{\LARGE \bf
Ada-Tracker: Soft Tissue Tracking via \\ Inter-Frame and Adaptive-Template Matching
}
\author{Jiaxin Guo$^{1}$,~\IEEEmembership{Student Member,~IEEE}, Jiangliu Wang$^{1}$,~\IEEEmembership{Member,~IEEE}, Zhaoshuo Li$^{2}$, Tongyu Jia$^{3}$, \\
Qi Dou$^{4,5}$,~\IEEEmembership{Member,~IEEE}, and Yun-Hui Liu$^{1,5}$, \textit{Fellow, IEEE}%
\thanks{This work is supported in part by Shenzhen Portion of Shenzhen-Hong Kong Science and Technology Innovation Cooperation Zone under HZQB-KCZYB-20200089, in part by the Research Grants Council of Hong Kong under Grant T42-409/18-R, Grant 14218322, and Grant 14207320, in part by the Hong Kong Centre for Logistics Robotics, in part by the Multi-Scale Medical Robotics Centre, InnoHK, and in part by the VC Fund 4930745 of the CUHK T Stone Robotics Institute. (\textit{Corresponding author: Yun-Hui Liu})}%
\thanks{$^{1}$CUHK T Stone Robotics Institute, The Chinese University of Hong Kong, Hong Kong.}
\thanks{$^{2}$Johns Hopkins University, United States.}
\thanks{$^{3}$Faculty of Urology, Third Medical Center, Chinese PLA General Hospital, Beijing, China.}
\thanks{$^{4}$Department of Computer Science and Engineering, The Chinese University of Hong Kong, Hong Kong.}
\thanks{$^{5}$Hong Kong Center for Logistics Robotics, Hong Kong.}
}

\begin{document}

\maketitle
\thispagestyle{empty}
\pagestyle{empty}

\begin{abstract}
Soft tissue tracking is crucial for computer-assisted interventions. Existing approaches mainly rely on extracting discriminative features from the template and videos to recover corresponding matches. However, it is difficult to adopt these techniques in surgical scenes, where tissues are changing in shape and appearance throughout the surgery. To address this problem, we exploit optical flow to naturally capture the pixel-wise tissue deformations and adaptively correct the tracked template. Specifically, we first implement an inter-frame matching mechanism to extract a coarse region of interest based on optical flow from consecutive frames. To accommodate appearance change and alleviate drift, we then propose an adaptive-template matching method, which updates the tracked template based on the reliability of the estimates. Our approach, Ada-Tracker, enjoys both short-term dynamics modeling by capturing local deformations and long-term dynamics modeling by introducing global temporal compensation. We evaluate our approach on the public SurgT benchmark, which is generated from Hamlyn, SCARED, and Kidney boundary datasets. The experimental results show that Ada-Tracker achieves superior accuracy and performs more robustly against prior works. Code is available at \url{https://github.com/wrld/Ada-Tracker}.

\end{abstract}

\section{INTRODUCTION}

Soft tissue tracking is an essential task in computer-assisted interventions, benefiting various downstream applications, including force estimation~\cite{giannarou2016vision}, motion compensation~\cite{richa2011towards}, image-guided surgery~\cite{yip2012tissue}, tissue scanning~\cite{wang2022towards, zhan2020autonomous}, autonomous tissue manipulation~\cite{wang2018unified},  3D Reconstruction~\cite{marmol2019dense, mahmoud2018live}, \etc. Tissue tracking also enhances tissue deformation estimation but also manipulation and interaction within surgical spaces~\cite{cartucho2023surgt}.

Typically, the soft tissue tracking technique involves selecting the target region of the tissue in the initial frame and then tracking the movement in subsequent frames of a surgical video sequence. Previous traditional methods employ rigid assumptions~\cite{grasa2011ekf, grasa2013visual, marmol2019dense}, classical descriptors~\cite{li2020super}, or model-based techniques~\cite{song2018mis, wong2012quasi, lurie20173d} to recover complex deformations on tissue surfaces. Recent studies~\cite{schmidt2022fast, schmidt2022recurrent, lin2023semantic} have revealed an emerging trend that utilizes data-driven techniques, which can be roughly divided into two categories: feature-based patch tracking and flow-based point tracking. 
\begin{figure}[t]
\centering
\includegraphics[width=1.0\linewidth]{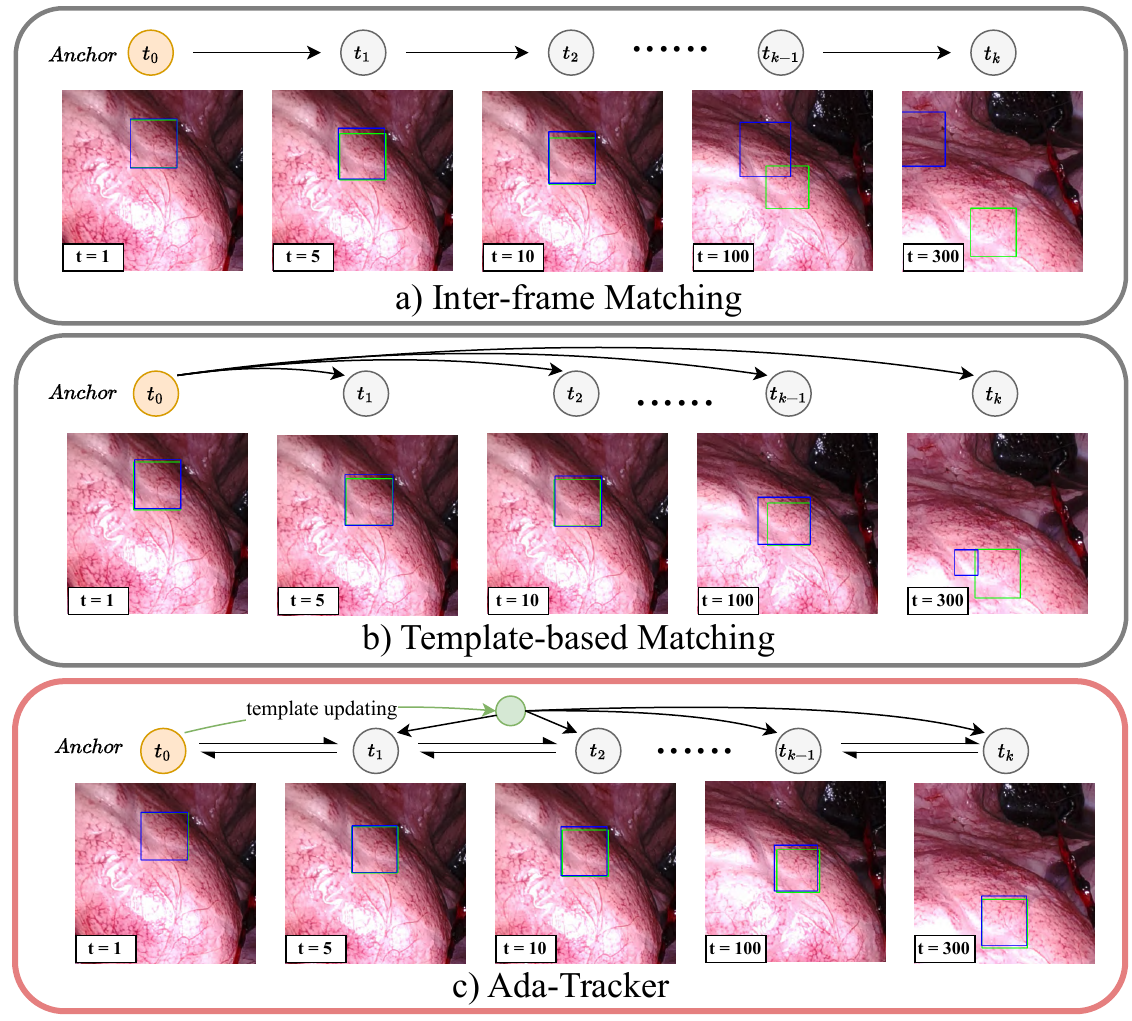}
\caption{\textbf{Different approaches to employ optical flow for soft tissue tracking,} with \textcolor{blue}{blue} and \textcolor{green}{green} bounding box indicating the predicted result and ground truth. a) Inter-frame matching updates the target box based on optical flow from consecutive frames, prone to drift due to accumulated error. b) Template-based matching predicts the target box by matching the template image from the anchor box with every input frame, struggling with appearance variation and large motion. c) Our Ada-Tracker exploits the benefits of a) and b) for the more robust and accurate tracker.}
\label{fig: teaser}
\vspace{-20pt}
\end{figure}
One major line of research is to consider the region of interest as an object and apply general visual object tracking (VOT) methods for tissue tracking~\cite{kristan2016novel, muller2018trackingnet, kristan2018sixth, huang2019got}. Such deep VOT trackers adopt a Siamese-based network~\cite{bertinetto2016fully} that independently extracts appearance features of the template and search region, then matches them via cross-correlation to determine the position with maximum similarity in the response map. However, surgical scenes possess a homogeneous appearance with texture-less soft tissue, making it difficult to correlate the template and search region to find the correspondence based on their appearance. %
To tackle this problem,  recent works attempt to estimate optical flow and capture the movements of individual points between neighboring frames for tissue tracking~\cite{neoral2023mft, doersch2022tap, schmidt2022recurrent}. Without relying on the appearance of objects during training, such flow-based methods offer better generalization to unseen deformable patterns, avoiding the challenges of adapting to different surgery domains. 
However, as they focus on estimating the motion of sparse points individually, such methods lack contextual information and are susceptible to noise and large deformations.

In this work, we argue that estimating the patch-level optical flow provides a comprehensive view of tissue deformation, enabling stable motion estimation with enhanced robustness. Tissue tracking based on patch-level optical flow could be more effective in modeling complex and non-rigid deformations, making it natural to handle complex deformations and occlusions. It motivates us to answer the following question: \textit{How to exploit patch-level optical flow for soft tissue tracking?} 
Two natural strategies to adapt patch-level optical flow for template-based soft tissue tracking are \textit{inter-frame matching} and \textit{template-based matching}. As shown in Fig. \ref{fig: teaser}, the first approach updates the target box by computing optical flow between consecutive frames. While this approach excels in achieving precise and stable motion estimation for short-term video sequences, it is prone to accumulate errors during long-term tracking, which can result in issues such as drifting, divergence, and sensitivity to occlusion. The template-based matching predicts the target region by taking the template image as the reference, to match with every frame. But this method struggles with challenges posed by variations in appearance and large movement, including fluctuations in illumination, camera motion, and deformations in soft tissue.

Based on the above analysis, we then propose Ada-Tracker for patch-level, long-term soft tissue tracking that exploits the strengths of both inter-frame motion estimation and template-based refinement, correlated with a designed adaptive template updating scheme. Ada-Tracker provides a comprehensive solution to soft tissue tracking by leveraging optical flow to capture both real-time tissue dynamics and exploit template-based correspondences. Specifically, our method consists of two stages: i) the \textbf{Inter-frame matching} stage estimates optical flow between consecutive frames, capturing the immediate motion of soft tissues to extract a coarse region of interest (ROI). We evaluate the confidence of the flow estimate to down-weigh outliers and check if occlusion is present. ii) the \textbf{Adaptive-template matching} stage then adaptively updates the template to reflect appearance changes while rectifying potential inaccuracies and drift from the initial inter-frame estimates. Ada-Tracker is capable of capturing the deformations of soft tissues with robustness and accuracy, countering surgical tracking challenges such as occlusions, drift, and appearance variations. 
Our contributions are summarized as threefold:
\begin{enumerate}
    \item We present a novel approach for soft tissue tracking by capitalizing on optical flow, offering a comprehensive solution that bridges optical flow with soft tissue tracking in surgical contexts.
    \item Our method harnesses the strengths of both inter-frame and adaptive-template matching, which estimates the soft tissue movement effectively, catering to both short-term changes and long-term trends.
    \item We perform thorough experiments to validate our method on the SurgT dataset, outperforming the previous SOTA trackers in terms of accuracy, and robustness in both 2D and 3D tissue tracking.
\end{enumerate}

\section{RELATED WORKS}
\noindent\textbf{Visual Object Tracking.}
Previous studies utilize CNN backbones~\cite{he2016deep},~\cite{krizhevsky2012imagenet} to extract features and fuse them through lightweight relation modeling networks, such as Siamese~\cite{bertinetto2016fully},~\cite{li2018high},~\cite{zhang2020ocean} and discriminative trackers~\cite{bhat2019learning},~\cite{danelljan2019atom}. 
However, their performance is restricted due to one-way information interaction. 
In response, methods like TransT~\cite{chen2021transformer} and STARK~\cite{yan2021learning} incorporate Transformers to allow bi-directional relation modeling, enhancing accuracy but at the cost of slower inference. 
By considering the region of interest as an object, some recent studies adopt the general VOT to address data-driven soft tissue tracking problem~\cite{cartucho2023surgt}. 
According to the report of the SurgT~\cite{cartucho2023surgt}, however, the low scores of TransT~\cite{chen2021transformer} indicate the difficulty of the generalization of the supervised learning-based method to the surgical scene. The VOT-based unsupervised tracking methods also struggle to find correspondence between the template and search region of homogeneous appearance, yielding lower performance compared to the traditional method.

\noindent\textbf{Soft Tissue Tracking.}
Soft tissue tracking is a non-rigid tracking method, that meets the challenges of texture-less, deformable tissue. Traditional works~\cite{grasa2011ekf, grasa2013visual, marmol2019dense} track the surgical scene with rigid assumptions, which fails when encountering large motions. MIS-SLAM~\cite{song2018mis} leverages deform nodes to model the surface deformations, and models like DefSLAM~\cite{lamarca2020defslam} combine meshes with classical features. To model the deformations, some work~\cite{wong2012quasi, lurie20173d} leverage spline and mesh to describe the deformations. Recent studies~\cite{schmidt2022fast, schmidt2022recurrent, lin2023semantic} have revealed an emerging trend that utilizes data-driven techniques using convolutional neural networks (CNN) or graph convolutional networks (GCN). However, a significant limitation of these efforts is their dependence on sparse key points or descriptors designed for specific datasets or applications. This restriction limits their usefulness in real-world dynamic surgical scenarios, which involve challenging situations such as instrument obstructions, changes in lighting, and significant distortions of soft tissue.

\begin{figure*}[t]
\centering
\includegraphics[width=0.95\linewidth]{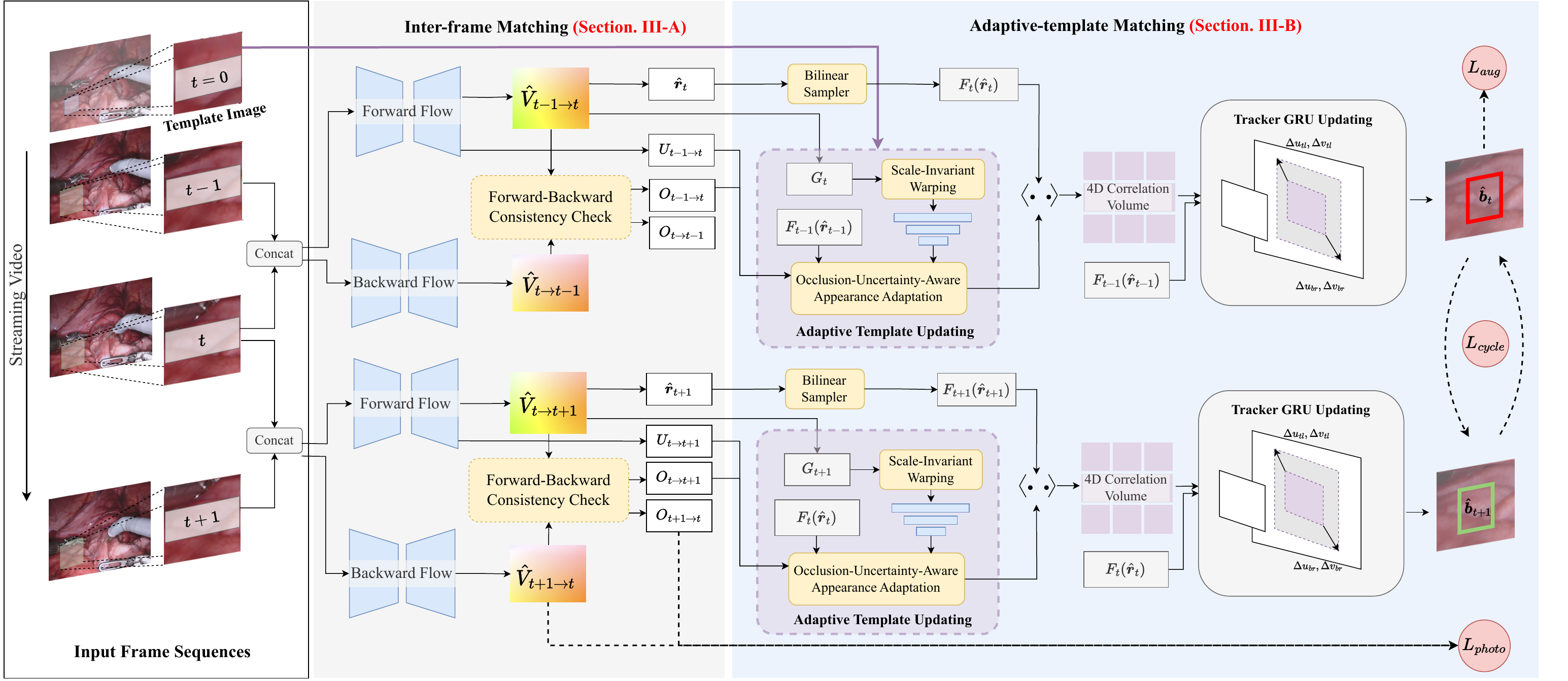}
\caption{\textbf{Overview of our proposed Ada-Tracker.} Given a bounding box at the start of the surgical video, we aim to locate the ROI of soft tissues in the form of bounding boxes at every frame. Our method consists of two stages: inter-frame matching and adaptive-template matching. In the first stage, we capture the immediate motion, and obtain a coarse bounding box prediction. Next, we update the template adaptively based on the flow, confidence and occlusion information from the previous stage. We finally match the updated template and coarse ROI to obtain the final prediction.}
\label{fig: method}
\vspace{-10pt}
\end{figure*}

\section{METHODOLOGY}
The overview of Ada-Tracker is illustrated in Fig. \ref{fig: method}. Ada-Tracker aims to locate the ROI of soft tissues in the form of bounding boxes at every frame, which is pre-defined at the start of a surgical video sequence. Our method follows a coarse-to-fine structure, consisting of two stages: 1) \textbf{Inter-frame matching} to capture the real-time dynamics of soft tissue movements, providing a coarse prediction for initialization (Sec. \ref{sec: local}), 2) \textbf{Adaptive-template matching} to first update the template image dynamically, and then refine the coarse prediction from the first stage. (Sec. \ref{sec: global})). We lastly introduce our self-supervised training and losses in Sec. \ref{sec: train}.

\subsection{Inter-frame Matching} \label{sec: local}
In the inter-frame matching stage, we aim to capture the immediate motion and obtain a coarse bounding box prediction of the tracked tissue. At the start of the surgical video, a bounding box $\bb_{0}$ at $I_{0}$ is assumed given, and the template patch $P_{0} \in \nR^{H_0 \times W_0 \times 3}$ within $\bb_{0}$ is extracted. At each time $t$, we initialize the search region in $I_t$ to be $P_{t} \in \nR^{4H_{t-1} \times 4W_{t-1} \times 3}$, which is expanded from the previous bounding box result $\hat{\bb}_{t-1}$. The tracked patch $P_{t-1} \in \nR^{4H_{t-1} \times 4W_{t-1} \times 3}$ from the previous image $I_{t-1}$ is also extracted by expanding $\hat{\bb}_{t-1}$. Both $P_{t}$ and $P_{t-1}$ are used to estimate a coarse bounding box at time $t$. 

\noindent\textbf{Uncertainty-aware Optical Flow Estimation} 
We leverage the RAFT model~\cite{teed2020raft} for our optical flow estimation to capture pixel-wise soft-tissue movement between consecutive frames $P_{t-1}$ and $P_{t}$.
Ada-Tracker begins with a feature encoder that transforms the input image pair $P_{t-1}$ and $P_{t}$ into a lower dimensional representation $F_{t-1}, F_{t} \in \nR ^{\frac{H_{t-1}}{8} \times \frac{W_{t-1}}{8} \times C}$ with $C=256$ to preserve the motion-critical details. $P_{t-1}$ is processed by the context encoder to provide spatial information during updating following~\cite{teed2020raft}. Subsequently, a 4D correlation volume is constructed, capturing the similarity between every pixel pair. A GRU-based updater is employed to iteratively refine optical flow predictions by exploiting both spatial and temporal contexts. 

To address the inherent uncertainty in motion estimation, we introduce an additional CNN layer to generate a confidence map \( U_{t-1 \rightarrow t} \) from the output of the GRU-based updater by a sigmoid activation. The inter-frame optical flow and its associated confidence are jointly computed as:
\begin{equation}
    \hat{V}_{t-1 \rightarrow t}, U_{t-1 \rightarrow t} = \mathcal{F}_{\theta} (I_{t-1},I_{t}),
\end{equation}
where \( \hat{V}_{t-1 \rightarrow t} \) denotes the estimated optical flow between the origin image patch \( I_{t-1} \) and the search image patch \( I_t \). The function \( \mathcal{F}_{\theta} \) refers to the inter-frame optical flow model parameterized by weights \( \theta \). The confidence map \( U_{t-1 \rightarrow t} \) quantifies the reliability of the predicted flows, highlighting regions with strong feature correspondences and thereby indicating areas of high stability. 

\noindent\textbf{Coarse ROI Prediction.} Based on the inter-frame optical flow $\hat{V}_{t-1 \rightarrow t}$, we update the coarse ROI $\hat{\br_t}$ from the previous coarse ROI $\hat{\br}_{t-1}$ which is double size expanded from the previous target bounding box $\hat{\bb}_{t-1}$. We combine the origin coordinates in $\hat{\br}_{t-1}$ with the predicted flow $\hat{V}_{t-1 \rightarrow t}(\hat{\br}_{t-1})$ as the predicted coordinates in $P_t$. Then $\hat{\br}_t$ is generated with a Min-Max strategy to extract the minimal enclosing rectangle of the predicted coordinates in $P_t$. 

\noindent\textbf{Occlusion Map.} To deal with the occlusion from instruments, camera movement, and tissue deformation, we apply the forward-backward consistency check following~\cite{meister2018unflow}. Specifically, we calculate the forward flow $\hat{V}_{t-1 \rightarrow t}$ and backward flow $\hat{V}_{t \rightarrow t-1}$, then generate the occlusion map $O_{t-1\rightarrow t}$ and $O_{t\rightarrow t-1}$. The occlusion map helps in identifying the occluded areas in the tracking region, serving as a mask to filter out the noisy feature with occlusion in the adaptive template updating.  During online tracking, we calculate the occlusion percentage $\text{occ}_{t-1 \rightarrow t}$ in $O_{t-1\rightarrow t}(\hat{\bb}_{t-1})$. No prediction will be made if occlusion percentage $\text{occ}_{t-1 \rightarrow t} > \beta$, where $\beta$ is an occlusion threshold.

\subsection{Adaptive-template Matching} \label{sec: global}
During long-term tissue tracking, matching a static template image with the search region is prone to fail when encountering drastic appearance change, deformation, and occlusion. To avoid the problem, the adaptive-template matching is introduced to dynamicly update the template image $P_0$ to improve the robustness under motion and appearance change. We update the template adaptively based on the flow, confidence, and occlusion information from the previous stage. Next, we devise an anchor-based matching network to find the best match of the updated template in the coarse ROI $\hat{\br_t}$. This anchor-based matching network follows a RAFT-like structure but focuses on patch-level updating instead of pixel-wise motion estimation, which is robust to noises and highly efficient.

\noindent\textbf{Adaptive Template Updating.} To aggregate the template image with spatial and temporal information from inter-frame dynamics, we enhance the anchor patch compensating both motion and appearance variation, from multi-aspect: the accumulated flow, occlusion map, and uncertainty map. 

For motion updating, we maintain the accumulated flow $G_{t}$, which contains the pixel-wise motion variation of the coarse target region from the start til now, \eg camera shift, scale-in, scale-out, and soft tissue deformation. To compensate for the motion variation, we design a scale-invariant warping scheme with the grid of the coarse ROI $\bx_0$, the center of the coarse ROI box $\bc_0$, the calculated scale ratio $S_t$:
\begin{equation}
\begin{aligned}
G_t &= \text{var}(G_{t-1} + V_{t-1 \rightarrow t}), \\
S_t &= \text{mean} (\sqrt{\frac{\parallel  -G_t + \bx_0 -\bc_0 \parallel^2}{\parallel \bx_0- 
\bc_0 \parallel^2}} ), \\
P_{w,t} &= P_0((-G_t + \bx_0 - \bc_0) \cdot S_t + \bc_0 ),
\end{aligned}
\end{equation}
where $P_{w, t}$ denotes the warped template image from the template image $P_0$. By using this scale-invariant warping approach, we could obtain the warped template image $P_{w,t}$ with inherent preservation of intricate details, such as rotations and tissue deformations, whilst simultaneously filtering out extreme alterations or positional shifts from camera-like scale-ins, scale-outs, or shifts. Then we extract warped template feature $F_{w, t}$ from $P_{w,t}$ by the feature encoder.

To address the appearance variation during the long-term surgical tracking, we resize the confidence map $U_{t-1 \rightarrow t}$ and occlusion map $O_{t-1 \rightarrow t}$ to match the size of $F_{t-1}$. Then $U_{t-1 \rightarrow t}$ and $O_{t-1 \rightarrow t}$ are leveraged to determine the weight for $F_{w, t}$ to adapt the appearance by fusing the target feature of previous step $F_{t-1}(\hat{\br}_{t-1})$. Our designed appearance adaptation is designed as follows:
\begin{equation}
\begin{aligned}
    F_{\text{occ}} &= (1 - O_{t \rightarrow t-1}(\hat{\br}_{t-1})) \odot F_{t-1}(\hat{\br}_{t-1}), \\
    F_{\text{conf}} &= U_{t-1 \rightarrow t}(\hat{\br}_{t-1}) \odot F_{\text{occ}}, \\
    F_{u, t} &= \alpha F_{w,t} + (1 - \alpha) F_{\text{conf}},
\end{aligned}
\end{equation}
where $F_{\text{occ}}$ and $F_{\text{conf}}$ denote the feature masked with the occlusion map,  and the feature with both uncertainty-awareness and occlusion-awareness, $F_{u, t}$ refers to the final updated template feature. Therefore, we ensure that occluded or boundary regions do not overly influence the fused feature representation.

\noindent\textbf{Template-based Matching.}
In this step, we aim to conduct an anchor-based matching between the updated template feature $F_{u, t} \in \nR ^{\frac{H_0}{8} \times \frac{W_0}{8}\times C}$ and the coarse ROI sampled search image feature $F_{t}(\hat{\br}_t) \in \nR^{\frac{H_{t-1}}{8} \times \frac{W_{t-1}}{8}\times C}$. Specifically, to correlate the template and search region, we build a 4D correlation cost volume $\bC \in \nR ^{\frac{H_0}{8} \times \frac{W_0}{8}\times  \frac{H_{t-1}}{8} \times \frac{W_{t-1}}{8}}$. We take the center half-size bounding box of $\hat{\br}_{t}$ as the initial index for the template feature to lookup the correlation volume, which enables a faster matching process based on the coarse initialization. Finally, we take $F_{t-1}(\hat{\br}_{t-1})$ as the context feature to serve the network with more spatial context around the template region.

Different from the pixel-wise motion updating in RAFT, our anchor-based GRU updates the region-wise motion to find the bounding box that best matches the search region. Specifically, our method only outputs the flow of the left-top corner and right-bottom corner to update the predicted bounding box. Then the updated flow is calculated as the interpolation of the corner flow:
\begin{equation}
     \hat{V}_{z}^{(i+1)} = \hat{V}_{z}^{(i)} + \text{Interp}(\Delta V_z^{(i)} (\bx_{tl}), \Delta V_z^{(i)} (\bx_{br})),
\end{equation}
where $\hat{V}_{z}^{(i)}$ refers to the predicted flow at iter $i$, $\Delta V_z^{(i)} (\bx_{tl})$ and $\Delta V_z^i (\bx_{br})$ denote the delta flow of the left top corner and bottom right corner. With the iterative updating of the template coordinates, we aim to match the template feature with the search region feature. Then the final predicted bounding box is calculated from the left top corner $\Delta V_z(\bx_{tl})$ and right bottom corner $\Delta V_z(\bx_{br})$ from the anchor-based matching model $\mathcal{F}_{\phi}$: 
\begin{equation}
\begin{aligned}
    \hat{V}_{z}, U_{z} &= \mathcal{F}_{\phi} (F_{u, t},F_{t}(\hat{\br}_t),F_{t-1}(\hat{\br}_{t-1}), \\
    \hat{\bb}_{t} &= \text{bbox}(\hat V_z(\bx_{tl}), \hat V_z(\bx_{br})),
\end{aligned}
\end{equation}
where $\hat{V}_z$ and $U_z$ indicate the tracking flow between the warped template image $F_{u, t}$ and the coarse search region $F_{t}(\hat{\br}_t)$. The designed anchor-based matching realizes a coarse-to-fine prediction of the tracking target, enabling improved tracking accuracy and robustness.

\subsection{Training and Losses} \label{sec: train}

\noindent\textbf{Cycle Consistency Loss.} As shown in Fig. \ref{fig: method}, after predicting the target bounding box $\hat{\bb}_{t-1 \rightarrow t}$ from $t-1$ to $t$, we crop the $I_{t+1}$ with four times expanded bounding box of the predicted result as the search region to track the anchor from $t$ to $t+1$. After this forward-tracking procedure, we take $\hat{\bb}_{t \rightarrow t+1}$ to start a backward-tracking from $t+1$ to $t-1$ to obtain the backward predicted result $\hat{\bb}_{cycle}$. After this cycle tracking, we could take the original anchor box $\bb$ as a pseudo label to supervise $\hat{\bb}_{cycle}$ by a linear combination of GIoU loss~\cite{rezatofighi2019generalized} and l1 norm loss. Additionally, we employ a cycle template loss to penalize the reconstruction loss between the warped template image $\hat{P}_{cycle}$ after the cycle and the template image $P$. Then our cycle loss is represented as follows:
\begin{equation}
\begin{aligned}
    \mathcal{L}_{cycle} = \mathcal{L}_{GIoU}(\hat{\bb}_{cycle}, \bb) &+ \mathcal{L}_{1}(\hat{\bb}_{cycle}, \bb) \\ &+ \mathcal{L}_{recon}(\hat{P}_{cycle}, P).
\end{aligned}
\end{equation}

\noindent\textbf{Total Loss.} To improve the self-supervise performance, we adopt an occlusion-aware photometric loss $\mathcal{L}_{photo}$~\cite{luo2021upflow} to train the inter-frame matching network, leveraging the occlusion map $O_{t-1 \rightarrow t}$ and $O_{t \rightarrow t-1}$. An augmentation loss $\mathcal{L}_{aug}$ is empolyed to supervise the tracking results with the pseudo ground truth label. A smooth loss $\mathcal{L}_{smooth}$~\cite{wang2018occlusion} is adopted to regularize the optical flow. Eventually, our training loss $\mathcal{L}$ is formulated as follows:
\begin{equation}
\mathcal{L} = \lambda_1 \mathcal{L}_{cycle} + 
\lambda_2 \mathcal{L}_{aug} + \lambda_3 \mathcal{L}_{photo} + \lambda_4 \mathcal{L}_{smooth},
\end{equation}
where $\lambda_1, \lambda_2, \lambda_3$, and $\lambda_4$ are loss weights.

 \section{EXPERIMENTS}  
\subsection{Datasets}
We evaluate our method in the recent proposed public SurgT dataset~\cite{cartucho2023surgt}. It provides the first standardized benchmark for assessing soft tissue tracking approaches and is generated from three datasets: Hamlyn~\cite{recasens2021endo}, SCARED~\cite{allan2021stereo}, and Kidney boundary datasets~\cite{hattab2020kidney}.  The SurgT dataset is composed of 157 stereo endoscopic videos with calibration parameters, including 125 videos from 12 cases for training, 12 videos from 3 cases for validation, and 20 videos from 5 cases for testing. Only the validation and testing datasets are annotated with the bounding boxes. Since no annotation is available for training, we randomly create the bounding box to train the Ada-tracker in a self-supervised manner. 

\begin{table}[t]
\caption{\textbf{Comparisons against prior works in the SurgT Challenge} on test set\label{table: test}}
\vspace{-3pt}
\renewcommand\theadfont{\setfsize}
\resizebox{\linewidth}{!}{%
\renewcommand{\arraystretch}{1.1}
\input{tables/comparison}
}
\vspace{-5mm}
\end{table}

\subsection{Implementation Details}
All experiments are implemented using PyTorch, and conducted on an NVIDIA RTX 3090 GPU. 
During training, we apply the AdamW optimizer with the learning rate as 0.000125 and the weight decay as 0.00001.
For loss weights, we set $\lambda_1 = 0.5, \lambda_2=0.1, \lambda_3=0.1, \lambda_4=0.001$. For training data, we collect video sequences from the training datasets in a random range, and random crop the $256 \times 256$ size images as input.  
We add shift, rotation, and illumination change to the image for augmentation and obtain the augmented bounding boxes as pseudo ground truth for $\mathcal{L}_{aug}$. Following SurgT~\cite{cartucho2023surgt}, we use both monocular metrics and stereo metrics to evaluate our method, including 2D/3D Accuracy, 2D/3D Error, 2D/2D Robustness, and Expected Average Overlap (EAO). 

\subsection{Comparison with State-of-the-art}
We compare Ada-Tracker with 7 methods submitted to the SurgT challenge, as well as CSRT~\cite{lukezic2017discriminative}, TransT~\cite{chen2021transformer} as two additional baselines. These methods can be grouped into three categories: i) traditional discriminative correlation filters-based tracker (CSRT~\cite{lukezic2017discriminative}, Jmees), ii) feature-based patch tracking method (TransT~\cite{chen2021transformer}, ETRI, MEDCVR, SRV, KIT, RIWOlink), and iii) optical flow based point tracking method (ICVS-2Ai).

\begin{figure}[t]
\centering
\includegraphics[width=1.0\linewidth]{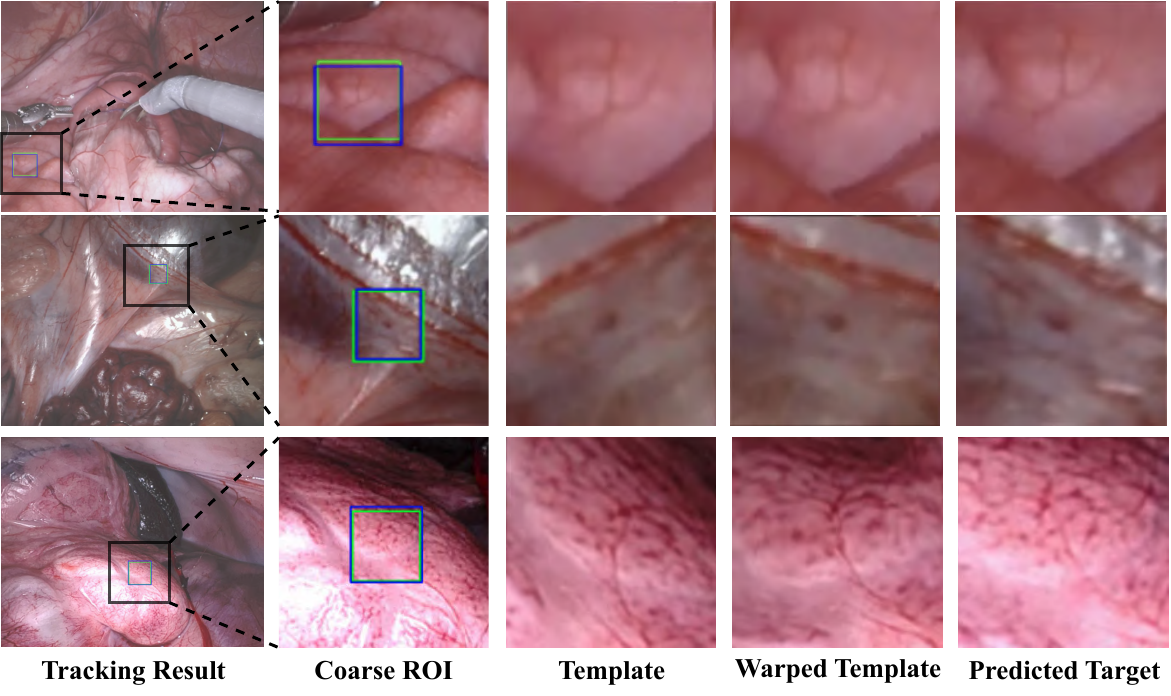}
\vspace{-6mm}
\caption{\textbf{Qualitative results}. We visualize the tracking in different cases, including tissue deformation, camera movement, and illumination variations.}
\label{fig: qual}
\vspace{-15pt}
\end{figure}
\begin{table*}[t]
\caption{\textbf{Comparisons against prior work in the SurgT Challenge} on test 2D cases \label{table: test2d}}
\vspace{-3pt}
\renewcommand\theadfont{\setfsize}
\resizebox{\linewidth}{!}{%
\renewcommand{\arraystretch}{1.1}
\input{tables/test_cases}
}
\vspace{-10pt}
\end{table*}

\begin{table*}[t]
\scriptsize
\caption{\textbf{Comparisons against prior work in the SurgT Challenge} on test 3D cases\label{table: test3d}}
\vspace{-3pt}
\renewcommand\theadfont{\setfsize}
\resizebox{\linewidth}{!}{%
\renewcommand{\arraystretch}{1.1}
\input{tables/test_case_3d}
}
\vspace{-10pt}
\end{table*}

As shown in Tab. \ref{table: test}, Ada-Tracker performs favorably against prior work in all metrics on case 1-5. The methods in the first category have achieved the most competitive performance, with Jmees achieving the best performance during the challenge. However, Jmees requires stereo frames to calculate the disparity information of consecutive frames to determine the size of the updated bounding boxes. They also require the pre-trained segmentation network to leverage prior tool information to determine the tracking occlusion. VOT-based methods leverage the generalization of models trained on the general computer vision datasets, which differ drastically from surgical scenes due to domain gaps. Regarding the optical flow-based ICVS-2Ai, while optical flow for point tracking improves the performance against appearance variation, their sensitivity to noises makes ICVS-Ai suffer from sudden scale change and occlusions during long-term tracking, as evidenced by the low robustness in Case 5 shown in Tab. \ref{table: test2d} and Tab. \ref{table: test3d}. Without patch-level optic flow to help determine the patch boundary, ICVS-Ai relies on left and right disparity to determine the predicted bounding box scale, which can affect the robustness in Tab. \ref{table: test3d}. 

On the contrary, by combining the benefits of inter-frame and adaptive-template matching, our Ada-Tracker addresses the limitations of both methods and performs more robustly than previous work. Our patch-level soft-tissue tracking tracks the target bounding box by estimating a holistic deformation of the template patch, overcoming surgical tracking challenges such as large tissue deformation (Case 3, 4), and random camera movements (Case 1, 2, 5), illumination and reflective surfaces (Case 1, 5). Without the use of stereo disparity information, our method is still capable of achieving competitive performance by utilizing scale-invariant warping in adaptive template updating. This maintains a constant scale reference point to assist in determining the target box boundary.  As shown in Fig. \ref{fig: qual}, while the current frame encounters camera rotation and illumination variation, our Ada-Tracker still achieves stable and accurate tracking, with the warped template remaining in the same motion state as the predicted target. More tracking visualization is provided in our supplementary video.

\begin{table}[t]
\small
\caption{\textbf{Ablation of Ada-Tracker} on validation set.\label{table: component} "I":inter-frame matching only, "T": template-based matching only, "A": adaptive template updating.}
\vspace{-3pt}
\renewcommand\theadfont{\setfsize}
\resizebox{\linewidth}{!}{%
\renewcommand{\arraystretch}{1.1}
\input{tables/ablation_components}
}
\vspace{-10pt}
\end{table}

\subsection{Ablation Study}
We evaluate the effectiveness of different components in our proposed method as shown in Tab. \ref{table: component}.

\paragraph{Inter-frame Matching} It updates the target box in the next frame based on the previous prediction and the estimated flow, resulting lowest robustness and accuracy in both 2D and 3D tracking. As shown in Fig. \ref{fig: qual}, when encountering large movements, the trajectory will get noise accumulation and eventually get a divergent trajectory.

\paragraph{Template-based Matching} Template-based matching estimates the target box based on finding the correspondences between the template image and the current frame, which improves the accuracy and robustness by setting a reference. It is shown in Fig. \ref{fig: qual} that in the first case, the trajectory of the template-based tracker could closely follow the ground truth trajectory most of the time. However, it eventually fails when the template appearance has a large difference with the search region.
\begin{figure}[ht!]
\centering
\includegraphics[width=0.95\linewidth]{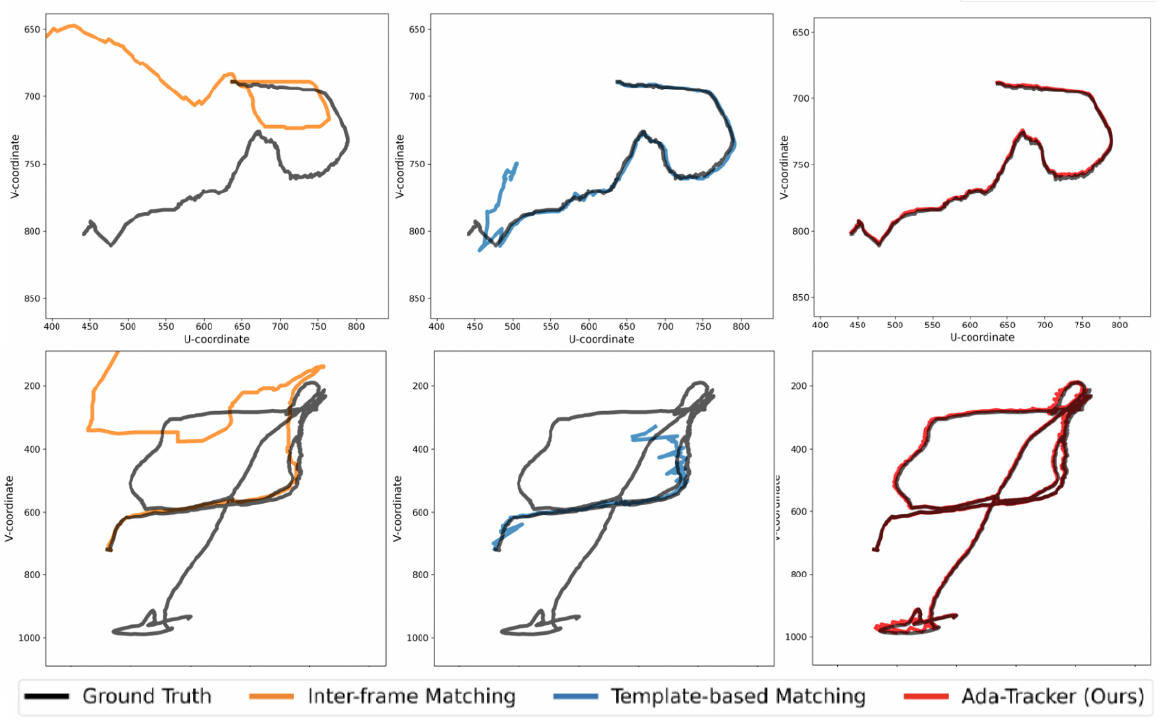}
\vspace{-3mm}
\caption{\textbf{Qualitative comparison} of different approaches in both short-term (first row) and long-term (second row) tracking case.}
\label{fig: traj}
\vspace{-5mm}
\end{figure}
\paragraph{Adaptive Template Updating} After adding the template updating scheme, we could improve the template image to make it consistent with the search region. As shown in Fig. \ref{fig: qual}, the warped template realizes consistent deformation and motion with the predicted target, making it easier to find correspondence for matching. The trajectory in Fig. \ref{fig: traj} shows that our trajectory closely matches the ground truth, verifying the effectiveness of our method.

\section{CONCLUSIONS}
In this paper, we propose Ada-Tracker to enable patch-level optical flow in soft tissue tracking, combining the strengths of both inter-frame and dynamic-template matching. Our method could identify movement from frame to frame and refine its accuracy through anchor-based modifications. Our Ada-Tracker achieves high robustness and accuracy encountering surgical obstacles like unexpected tissue movements or appearance alterations. 
Our work has the great potential to be applied to computer-assisted interventions, helping make surgeries safer and more streamlined.

\bibliographystyle{IEEEtran}
\bibliography{egbib}
\end{document}

%% file: shortcut.tex
\newcommand{\bb}{\mathbf{b}}
\newcommand{\bc}{\mathbf{c}}\newcommand{\bC}{\mathbf{C}}

\newcommand{\br}{\mathbf{r}}

\newcommand{\bx}{\mathbf{x}}

\newcommand{\nR}{\mathbb{R}}

\makeatletter
\DeclareRobustCommand\onedot{\futurelet\@let@token\@onedot}
\def\@onedot{\ifx\@let@token.\else.\null\fi\xspace}
\def\eg{e.g\onedot}

\def\etc{etc\onedot}

\makeatother

%% file: tables/comparison.tex
\begin{tabular}{@{}C{1.2cm}|C{0.7cm}C{0.7cm}C{1.0cm}C{0.7cm}C{0.7cm}C{1.0cm}@{}}
\toprule
Method  & $Rob_{2D}$$\uparrow$ & $Acc_{2D}$$\uparrow$ & $Err_{2D}$$\downarrow$ & $Rob_{3D}$$\uparrow$ & $Err_{3D}$$\downarrow$&$EAO$$\uparrow$ 
\\\cmidrule(l){1-7}
CSRT&\underline{0.872}&0.769&5.7$\pm$2.6&0.894&3.3$\pm$3.8&0.563\\
Jmees&0.868&\underline{0.818}&\underline{5.3$\pm$2.4}&0.878&2.7$\pm$1.9&\underline{0.583} \\
KIT& 0.465  &   0.747  & 12.5$\pm$8.0& 0.810 &  11.9$\pm$9.5 & 0.223  \\
TransT& 0.701 & 0.529 &   16.3$\pm$5.1 & 0.861&   17.3$\pm$26.7 & 0.274 \\
SRV& 0.476 &   0.681 & 15.4$\pm$7.5 &   0.710 &8.5$\pm$13.7&   0.293  \\
MEDCVR & 0.702 & 0.509 &  7.9$\pm$5.8 & 0.832&   9.2$\pm$39.7 & 0.302 \\
ETRI & 0.802 &  0.693 & 12.1$\pm$7.4 & \underline{0.909} & 5.7$\pm$5.2 & 0.405  \\
RIWOlink &0.807&0.737&8.0$\pm$4.5&0.894&5.8$\pm$9.2&0.433 \\

ICVS-2Ai & \underline{0.872} &0.816&6.7$\pm$3.5&0.901&\underline{2.6$\pm$2.3}&0.573 \\
\textbf{Ours} & \textbf{0.894} & \textbf{0.833} & \textbf{5.2$\pm$3.3} & \textbf{0.912} & \textbf{2.1$\pm$2.0} & \textbf{0.591} \\
\hline 
\end{tabular}

%% file: tables/test_cases.tex
\begin{tabular}{@{}C{1.2cm}|C{0.7cm}C{0.7cm}C{1.0cm}|C{0.7cm}C{0.7cm}C{1.0cm}|C{0.7cm}C{0.7cm}C{1.0cm}|C{0.7cm}C{0.7cm}C{1.0cm}|C{0.7cm}C{0.7cm}C{1.0cm}@{}}
\toprule
 \multicolumn{1}{c}{\multirow{2}{*}{\makecell{Method}}}&  \multicolumn{3}{c}{\multirow{1}{*}{\makecell{Case 1}}} & \multicolumn{3}{c}{\multirow{1}{*}{\makecell{Case 2}}} &\multicolumn{3}{c}{\multirow{1}{*}{\makecell{Case 3}}} &\multicolumn{3}{c}{\multirow{1}{*}{\makecell{Case 4}}} & \multicolumn{3}{c}{\multirow{1}{*}{\makecell{Case 5}}}\\\cmidrule(l){2-16} 
\multicolumn{1}{c}{}  & $Rob_{2D}$$\uparrow$ & $Acc_{2D}$$\uparrow$ & $Err_{2D}$$\downarrow$& $Rob_{2D}$$\uparrow$ & $Acc_{2D}$$\uparrow$ & $Err_{2D}$$\downarrow$ & $Rob_{2D}$$\uparrow$ & $Acc_{2D}$$\uparrow$ & $Err_{2D}$$\downarrow$&$Rob_{2D}$$\uparrow$ & $Acc_{2D}$$\uparrow$ & $Err_{2D}$$\downarrow$&$Rob_{2D}$$\uparrow$ & $Acc_{2D}$$\uparrow$ & $Err_{2D}$$\downarrow$
\\\cmidrule(l){1-16}
CSRT&\multicolumn{1}{c|}{0.92} & \multicolumn{1}{c|}{0.72} & \multicolumn{1}{c|}{\textbf{7±3}} & \multicolumn{1}{c|}{0.87} & \multicolumn{1}{c|}{0.77} & \multicolumn{1}{c|}{8±3} & \multicolumn{1}{c|}{0.87} & \multicolumn{1}{c|}{0.81} & \multicolumn{1}{c|}{5±2} & \multicolumn{1}{c|}{\textbf{1.0}} & \multicolumn{1}{c|}{0.85} & \multicolumn{1}{c|}{\textbf{3±1}} & \multicolumn{1}{c|}{\underline{0.70}} & \multicolumn{1}{c|}{0.68} & \multicolumn{1}{c}{6±3} \\
Jmees&\multicolumn{1}{c|}{0.92} & \multicolumn{1}{c|}{\underline{0.76}} & \multicolumn{1}{c|}{\textbf{7±3}} & \multicolumn{1}{c|}{\textbf{0.89}} & \multicolumn{1}{c|}{0.77} & \multicolumn{1}{c|}{\textbf{7±3}} & \multicolumn{1}{c|}{\underline{0.88}} & \multicolumn{1}{c|}{\textbf{0.88}} & \multicolumn{1}{c|}{5±2} & \multicolumn{1}{c|}{0.94} & \multicolumn{1}{c|}{\textbf{0.90}} & \multicolumn{1}{c|}{\textbf{3±1}} & \multicolumn{1}{c|}{\underline{0.70}} & \multicolumn{1}{c|}{\underline{0.75}} & \multicolumn{1}{c}{\underline{5±3}}  \\
KIT&  \multicolumn{1}{c|}{0.26} & \multicolumn{1}{c|}{0.61} & \multicolumn{1}{c|}{19±13}    & \multicolumn{1}{c|}{0.17} & \multicolumn{1}{c|}{0.61} & \multicolumn{1}{c|}{23±17}    & \multicolumn{1}{c|}{0.81} & \multicolumn{1}{c|}{0.84} & \multicolumn{1}{c|}{9±7}      & \multicolumn{1}{c|}{0.84} & \multicolumn{1}{c|}{0.71} & \multicolumn{1}{c|}{11±6}     & \multicolumn{1}{c|}{0.16} & \multicolumn{1}{c|}{0.58} & \multicolumn{1}{c}{20±10} \\
TransT&\multicolumn{1}{c|}{0.72} & \multicolumn{1}{c|}{0.46} & \multicolumn{1}{c|}{18±4}    & \multicolumn{1}{c|}{0.64} & \multicolumn{1}{c|}{0.42} & \multicolumn{1}{c|}{24±10}    & \multicolumn{1}{c|}{0.82} & \multicolumn{1}{c|}{0.61} & \multicolumn{1}{c|}{16±4}      & \multicolumn{1}{c|}{0.79} & \multicolumn{1}{c|}{0.61} & \multicolumn{1}{c|}{11±7}     & \multicolumn{1}{c|}{0.47} & \multicolumn{1}{c|}{0.48} & \multicolumn{1}{c}{11±4}    \\
SRV& \multicolumn{1}{c|}{0.39} & \multicolumn{1}{c|}{0.49} & \multicolumn{1}{c|}{27±13}    & \multicolumn{1}{c|}{0.19} & \multicolumn{1}{c|}{0.55} & \multicolumn{1}{c|}{16±11}    & \multicolumn{1}{c|}{0.71} & \multicolumn{1}{c|}{0.80} & \multicolumn{1}{c|}{11±6}     & \multicolumn{1}{c|}{0.98} & \multicolumn{1}{c|}{0.69} & \multicolumn{1}{c|}{13±5}     & \multicolumn{1}{c|}{0.07} & \multicolumn{1}{c|}{0.56} & \multicolumn{1}{c}{22±12}     \\
MEDCVR & \multicolumn{1}{c|}{0.74} & \multicolumn{1}{c|}{0.54} & \multicolumn{1}{c|}{\underline{7±5}}      & \multicolumn{1}{c|}{0.65} & \multicolumn{1}{c|}{0.48} & \multicolumn{1}{c|}{16±12}    & \multicolumn{1}{c|}{0.82} & \multicolumn{1}{c|}{0.68} & \multicolumn{1}{c|}{4±4}      & \multicolumn{1}{c|}{0.77} & \multicolumn{1}{c|}{0.61} & \multicolumn{1}{c|}{10±7}     & \multicolumn{1}{c|}{0.46} & \multicolumn{1}{c|}{0.57} & \multicolumn{1}{c}{6±4}       \\
ETRI &  \multicolumn{1}{c|}{0.88} & \multicolumn{1}{c|}{0.56} & \multicolumn{1}{c|}{21±13}    & \multicolumn{1}{c|}{0.82} & \multicolumn{1}{c|}{0.69} & \multicolumn{1}{c|}{14±10}    & \multicolumn{1}{c|}{0.85} & \multicolumn{1}{c|}{0.82} & \multicolumn{1}{c|}{7±3}      & \multicolumn{1}{c|}{0.94} & \multicolumn{1}{c|}{0.73} & \multicolumn{1}{c|}{\underline{7±4}}      & \multicolumn{1}{c|}{0.49} & \multicolumn{1}{c|}{0.60} & \multicolumn{1}{c}{10±6}  \\
RIWOlink & \multicolumn{1}{c|}{0.89} & \multicolumn{1}{c|}{0.66} & \multicolumn{1}{c|}{9±5}      & \multicolumn{1}{c|}{0.71} & \multicolumn{1}{c|}{0.78} & \multicolumn{1}{c|}{9±6}      & \multicolumn{1}{c|}{0.85} & \multicolumn{1}{c|}{0.86} & \multicolumn{1}{c|}{\textbf{3±1}}      & \multicolumn{1}{c|}{\underline{0.99}} & \multicolumn{1}{c|}{0.69} & \multicolumn{1}{c|}{10±5}     & \multicolumn{1}{c|}{0.55} & \multicolumn{1}{c|}{0.59} & \multicolumn{1}{c}{9±6}  \\

ICVS-2Ai & \multicolumn{1}{c|}{\underline{0.94}} & \multicolumn{1}{c|}{\underline{0.76}} & \multicolumn{1}{c|}{9±4} & \multicolumn{1}{c|}{0.84} & \multicolumn{1}{c|}{0.79} & \multicolumn{1}{c|}{9±6} & \multicolumn{1}{c|}{\textbf{0.91}} & \multicolumn{1}{c|}{\textbf{0.88}} & \multicolumn{1}{c|}{5±3} & \multicolumn{1}{c|}{\textbf{1.0}} & \multicolumn{1}{c|}{\textbf{0.90}} & \multicolumn{1}{c|}{\textbf{3±1}} & \multicolumn{1}{c|}{0.65} & \multicolumn{1}{c|}{0.72} & \multicolumn{1}{c}{8±5}  \\
\textbf{Ours} & \multicolumn{1}{c|}{\textbf{0.95}} & \multicolumn{1}{c|}{\textbf{0.79}} & \multicolumn{1}{c|}{\underline{7±5}} & \multicolumn{1}{c|}{\underline{0.88}} & \multicolumn{1}{c|}{\textbf{0.82}} & \multicolumn{1}{c|}{\underline{7±6}} & \multicolumn{1}{c|}{\textbf{0.91}} & \multicolumn{1}{c|}{\textbf{0.88}} & \multicolumn{1}{c|}{\underline{4±3}} & \multicolumn{1}{c|}{\textbf{1.0}} & \multicolumn{1}{c|}{\underline{0.88}} & \multicolumn{1}{c|}{\textbf{3±1}} & \multicolumn{1}{c|}{\textbf{0.72}} & \multicolumn{1}{c|}{\textbf{0.78}} & \multicolumn{1}{c}{\textbf{4±3}}  \\
\hline 
\end{tabular}

%% file: tables/test_case_3d.tex
\setlength\tabcolsep{1.4em}
\begin{tabular}{@{}C{1.2cm}|C{0.7cm}C{1.0cm}|C{0.7cm}C{1.0cm}|C{0.7cm}C{1.0cm}|C{0.7cm}C{1.0cm}|C{0.7cm}C{1.0cm}@{}}
\toprule
 \multicolumn{1}{c}{\multirow{2}{*}{\makecell{Method}}}&  \multicolumn{2}{c}{\multirow{1}{*}{\makecell{Case 1}}} & \multicolumn{2}{c}{\multirow{1}{*}{\makecell{Case 2}}} &\multicolumn{2}{c}{\multirow{1}{*}{\makecell{Case 3}}} &\multicolumn{2}{c}{\multirow{1}{*}{\makecell{Case 4}}} & \multicolumn{2}{c}{\multirow{1}{*}{\makecell{Case 5}}}\\\cmidrule(l){2-11} 
\multicolumn{1}{c}{}  & $Rob_{3D}$$\uparrow$ &  $Err_{3D}$$\downarrow$& $Rob_{3D}$$\uparrow$ &  $Err_{3D}$$\downarrow$ & $Rob_{3D}$$\uparrow$ & $Err_{3D}$$\downarrow$&$Rob_{3D}$$\uparrow$& $Err_{3D}$$\downarrow$& $Rob_{3D}$$\uparrow$ &  $Err_{3D}$$\downarrow$ 
\\\cmidrule(l){1-11}
CSRT&\multicolumn{1}{c|}{0.91}  &\multicolumn{1}{c|}{5±4}& \multicolumn{1}{c|}{0.83} &\multicolumn{1}{c|}{\underline{2±1}}& \multicolumn{1}{c|}{0.93} &\multicolumn{1}{c|}{\textbf{1±1}}& \multicolumn{1}{c|}{\textbf{1.0}} &\multicolumn{1}{c|}{\textbf{1±1}}& \multicolumn{1}{c|}{\textbf{0.78}} &\multicolumn{1}{c}{9±16} \\
Jmees& \multicolumn{1}{c|}{\textbf{1.0}}  & \multicolumn{1}{c|}{\textbf{2±1}}& \multicolumn{1}{c|}{\underline{0.85}} & \multicolumn{1}{c|}{\underline{2±1}}& \multicolumn{1}{c|}{0.93} & \multicolumn{1}{c|}{\textbf{1±1}}& \multicolumn{1}{c|}{0.95} & \multicolumn{1}{c|}{\textbf{1±1}}& \multicolumn{1}{c|}{0.72} & \multicolumn{1}{c}{\underline{5±4}}   \\
KIT& \multicolumn{1}{c|}{0.68} &\multicolumn{1}{c|}{20±16}   & \multicolumn{1}{c|}{0.70} &\multicolumn{1}{c|}{23±14}   & \multicolumn{1}{c|}{\textbf{0.97}} &\multicolumn{1}{c|}{2±2}& \multicolumn{1}{c|}{\textbf{1.0}}  &\multicolumn{1}{c|}{6±5}& \multicolumn{1}{c|}{0.68} &\multicolumn{1}{c}{20±19}   \\
TransT&\multicolumn{1}{c|}{0.91} &\multicolumn{1}{c|}{23±8}   & \multicolumn{1}{c|}{0.82} &\multicolumn{1}{c|}{24±53}   & \multicolumn{1}{c|}{0.92} &\multicolumn{1}{c|}{6±14}& \multicolumn{1}{c|}{0.92}  &\multicolumn{1}{c|}{10±56}& \multicolumn{1}{c|}{0.70} &\multicolumn{1}{c}{31±26}    \\
SRV&\multicolumn{1}{c|}{0.66} &\multicolumn{1}{c|}{16±13}   & \multicolumn{1}{c|}{0.37} &\multicolumn{1}{c|}{16±12}   & \multicolumn{1}{c|}{\underline{0.96}} &\multicolumn{1}{c|}{3±3}& \multicolumn{1}{c|}{\textbf{1.0}}  &\multicolumn{1}{c|}{\underline{2±1}}& \multicolumn{1}{c|}{0.46} &\multicolumn{1}{c}{18±66}  \\
MEDCVR & \multicolumn{1}{c|}{0.90} & \multicolumn{1}{c|}{8±22}    & \multicolumn{1}{c|}{0.75} & \multicolumn{1}{c|}{20±125}  & \multicolumn{1}{c|}{0.88} & \multicolumn{1}{c|}{2±9}& \multicolumn{1}{c|}{0.92} & \multicolumn{1}{c|}{10±62}   & \multicolumn{1}{c|}{0.67} & \multicolumn{1}{c}{14±24}  \\
ETRI &  \multicolumn{1}{c|}{0.97} & \multicolumn{1}{c|}{6±4}& \multicolumn{1}{c|}{\textbf{0.90}} & \multicolumn{1}{c|}{7±6}& \multicolumn{1}{c|}{0.94} & \multicolumn{1}{c|}{2±3}& \multicolumn{1}{c|}{\textbf{1.0}}  & \multicolumn{1}{c|}{3±3}& \multicolumn{1}{c|}{0.71} & \multicolumn{1}{c}{13±13}   \\
RIWOlink & \multicolumn{1}{c|}{0.96} & \multicolumn{1}{c|}{7±4}& \multicolumn{1}{c|}{0.81} & \multicolumn{1}{c|}{6±5}& \multicolumn{1}{c|}{0.93} & \multicolumn{1}{c|}{\textbf{1±1}}& \multicolumn{1}{c|}{\underline{0.99}} & \multicolumn{1}{c|}{5±34}    & \multicolumn{1}{c|}{0.72} & \multicolumn{1}{c}{11±7}    
 \\
ICVS-2Ai &\multicolumn{1}{c|}{0.98}  & \multicolumn{1}{c|}{4±2}& \multicolumn{1}{c|}{0.84} & \multicolumn{1}{c|}{\textbf{1±1}}& \multicolumn{1}{c|}{0.94} & \multicolumn{1}{c|}{\underline{1±2}}& \multicolumn{1}{c|}{\textbf{1.0}} & \multicolumn{1}{c|}{\textbf{1±1}}& \multicolumn{1}{c|}{0.71} & \multicolumn{1}{c}{6±6}  \\

Ours &\multicolumn{1}{c|}{\underline{0.99}}  & \multicolumn{1}{c|}{\underline{2±2}}& \multicolumn{1}{c|}{0.84} & \multicolumn{1}{c|}{\underline{2±1}}& \multicolumn{1}{c|}{\underline{0.96}} & \multicolumn{1}{c|}{\textbf{1±1}}& \multicolumn{1}{c|}{\textbf{1.0}} & \multicolumn{1}{c|}{\textbf{1±1}}& \multicolumn{1}{c|}{\underline{0.76}} & \multicolumn{1}{c}{\textbf{4±3}}  \\
\hline 
\end{tabular}

%% file: tables/ablation_components.tex
\begin{tabular}{@{}C{0.3cm}C{0.3cm}C{0.3cm}|C{1cm}C{1cm}C{1cm}C{1cm}C{1cm}C{1cm}@{}}
\toprule
I & T & A &$Rob_{2D}$$\uparrow$ & $Acc_{2D}$$\uparrow$ & $Err_{2D}$$\downarrow$ & $Rob_{3D}$$\uparrow$ & $Err_{3D}$$\downarrow$&$EAO$$\uparrow$
\\\cmidrule(l){1-9}
 \checkmark &   &     & 0.323 & 0.512  &14.2$\pm$5.4& 0.587 & 16.9$\pm$13.5 & 0.088\\
  &  \checkmark  &  & 0.541 & 0.627  &8.9$\pm$7.3& 0.620 & 9.8$\pm$19.7 & 0.164 \\
  &  \checkmark  &\checkmark  & 0.643 & 0.639  &7.2$\pm$6.5& 0.693 & 7.9$\pm$17.4 & 0.279 \\
  \checkmark & \checkmark &\checkmark   & \textbf{0.740} & \textbf{0.845}  &\textbf{4.2$\pm$5.6}& \textbf{0.783} & \textbf{2.8$\pm$2.5} & \textbf{0.318} \\ \hline 
\end{tabular}